\documentclass{article}
\usepackage[utf8]{inputenc}
\usepackage{authblk}
\usepackage{xcolor}
\usepackage{graphicx}
\usepackage{booktabs}       
\usepackage{amsmath}
\usepackage{mathtools}
\usepackage{gensymb}
\usepackage{booktabs}       
\usepackage{amsfonts}       
\usepackage{nicefrac}       
\usepackage{microtype}      
\usepackage{algorithm}
\usepackage{algorithmic}
\usepackage{mathtools}
\usepackage[
    natbib=true,
    style=nature,
    sorting=none
]{biblatex}
\usepackage{hyperref}
\hypersetup{breaklinks=true}
\bibliography{main}
\newcommand{\Figure}[1]{Fig.~\ref{#1}}

\newcommand{\bs}{\boldsymbol}

\makeatletter
\newcommand{\mypm}{\mathbin{\mathpalette\@mypm\relax}}
\newcommand{\@mypm}[2]{\ooalign{%
  \raisebox{.1\height}{$#1+$}\cr
  \smash{\raisebox{-.6\height}{$#1-$}}\cr}}
\makeatother
\usepackage{graphicx}
\title{Personalized Pose Forecasting}
\author[1]{Maria  Priisalu}
\author[1]{Ted Kronvall}
\author[1,2]{Cristian Sminchisescu}
\affil[1]{Center of Mathematics, Faculty of Eningeering, Lund University}
\affil[2]{Google Research}

\begin{document}
\maketitle

\begin{abstract}
    Human pose forecasting is the task of predicting articulated human motion given past human motion. There exists a number of popular benchmarks that evaluate an array of different models performing human pose forecasting. These benchmarks do not reflect that a human interacting system, such as a delivery robot, observes and plans for the motion of the same individual over an extended period of time. Every individual has unique and distinct movement patterns. This is however not reflected in existing benchmarks that evaluate a model's ability to predict an average human's motion rather than a particular individual's. We reformulate the human motion forecasting problem and present a model-agnostic personalization method. Motion forecasting personalization can be performed efficiently online by utilizing a low-parametric time-series analysis model that personalizes neural network pose predictions. 
\end{abstract}
\pagestyle{empty}
\section{Introduction}
The number of  artificial intelligence (AI) systems that interact with humans is increasing, from autonomous vehicles to industrial robots, and virtual reality systems. Human-Computer Interaction (HCI) however requires the AI system to perceive humans, and in many cases, the AI system must also forecast the human's future motion (to avoid collisions between the human and a robot or to plan for the human's future motion).  A typical HCI system will interact with an individual for some time period. For example, an autonomous package delivery robot will need to pass various individuals on the street. The interaction occurs while the human is visible to the robot, varying in time from a number of seconds to a number of minutes. We argue that a typical HCI is long enough for the system to adapt its human motion prediction to the specific individual, and propose a lightweight online method to do so.

Each individual has varying limb lengths, joint agility, and person-specific behavior traits. Therefore it is natural that the motion of humans varies from one individual to another. Each individual has specific motion patterns that they tend to adhere to. This is utilized in human gait recognition~\cite{wan2018survey}. Human gait recognition is the task of recognizing an individual from their walking motion pattern alone, allowing for individual recognition from a distance for example from surveillance cameras. These individual-specific motion patterns could be used to improve individual-specific articulated human motion prediction. 

 Human pose, that is the locations of selected joints in the human skeleton, is a natural and compact representation of humans. There exists a wide variety of human pose forecasting models that are evaluated on the popular benchmark datasets Human3.6M~\cite{ionescu2013human3}, HumanEva~\cite{sigal2010humaneva}, and CMU Motion Capture Dataset~\cite{CMUMoCaP}. The existing benchmarks~\cite{ionescu2013human3,sigal2010humaneva,CMUMoCaP} are motivated by human motion animation and concentrate on forecasting an average human's motion for a short time window typically $1s$ given past $0.4s$ of history. This is a very short observation window and in practice, an AI system (such as an autonomous package delivery robot) is likely to have longer interactions with a specific individual. Therefore we argue that for practical implementations it is of greater interest to evaluate the various human pose forecasting models on their prediction ability of a specific individual given all available data. 

Existing human pose forecasting models should be adapted to individuals online, such that as a system interaction with an individual becomes longer in time the model predictions improve. However, existing Neural Network(NN) based approaches are still parameter-heavy, so updating parameters online in the presence of possibly limited resources is hard. Personalizing NN-based approaches is not easy as caution must be taken to avoid catastrophic forgetting~\cite{mccloskey1989catastrophic}. Observations of the joint motion of an individual will naturally be correlated, leading to high variance estimates of the gradient during training. This must also be treated with care. Finally providing a longer time-horizon observation to NNs to allow the utilization of temporal patterns of an individual does not guarantee increased performance of the NN on the prediction task. To avoid the temperamental behavior of NNs during finetuning an NN to a specific individual we use individual-specific correcting low parametric models. Low-parametric models can be trained online to fit any new individuals or new motion patterns.

We show that Autoregressive (AR) models~\cite{box2015time} from time-series analysis  with a few hundred parameters have a performance that is comparable to neural networks-based methods with $\approx$0.14-16M parameters. 
With only a few hundred parameters classical time-series analysis methods can be adapted to the data of a particular individual on the fly. We take a first step in personalizing human motion forecasting by utilizing an AR model to fine-tune the predictions of NNs to the tested individual during test time and study how well state-of-the-art models capture personalized motion. Applying an AR model on the network residuals requires no intervention with the model or re-training of the NN, and provides a model-agnostic approach to personalizing predictions. This is only a first step because AR models are low parametric and well suited to model temporal dependence and can adapt any neural predictions while being updated online, but have an exponentially decreasing prediction power. To personalize long-term predictions model-specific personalization is needed, this is left as future work. The analysis is performed on a few selected models. 

\section{Related work}
Techniques to capture human joint motion have improved~\cite{tallamraju2020aircaprl,palermo2022raw,von2018recovering,shukla2022vl4pose,AMASS:ICCV:2019}, allowing for approximate human motion capture in the wild~\cite{von2018recovering}, and articulated datasets for human-robot interaction~\cite{sampieri2022pose}. Nonetheless, Human3.6M~\cite{ionescu2013human3} continues to be the most popular dataset~\cite{lyu20223d} to benchmark human pose forecasting methods. To provide insight into the capabilities of numerous existing models already benchmarked on the most popular dataset we propose to adjust the forecasting task of Human3.6M~\cite{ionescu2013human3} and propose a general method to personalize human forecasting models.

A number of improvements of the popularly benchmarked human pose forecasting problem on datasets~\cite{ionescu2013human3,sigal2010humaneva,CMUMoCaP} have been suggested in the literature; a longer prediction time horizon (up to 5s) has been proposed in~\cite{kiciroglu2022keyposes,ma2022progressively,diller2022charposes,zang2022few}, probabilistic pose forecasting in~\cite{xu2021probabilistic,ding2021uncertainty,saadatnejad2022generic}, evaluation in the presence of occlusions in~\cite{drumond2022few,adeli2021tripod}, prediction when missing data in~\cite{kieu2022locally}, generalization to new actions in~\cite{Gui2018FewShotHM}, and jointly predicting human trajectory and pose ~\cite{nikdel2022dmmgan,adeli2021tripod,adeli2020socially}.
The existing benchmarks on~\cite{ionescu2013human3,sigal2010humaneva,CMUMoCaP} are evaluated for an average human. We propose to personalize human motion forecasting benchmarking. Personalized human pose forecasting is partially motivated by the success of personalized human trajectory forecasting~\cite{zhu2022personalized} and by the success of gait recognition~\cite{wan2018survey,balazia2018gait,singh2018vision}. The unexpectedly good performance of parameter sparse models in human motion forecasting has also been noted in previous work~\cite{martinez2017human,guo2023back}. 

There exists a large amount of human forecasting methods~\cite{lyu20223d,sampieri2022pose,kiciroglu2022keyposes,ma2022progressively,diller2022charposes,zang2022few,xu2021probabilistic,ding2021uncertainty,saadatnejad2022generic,drumond2022few,kieu2022locally,Gui2018FewShotHM,nikdel2022dmmgan,adeli2021tripod,adeli2020socially,zhong2022spatial,starke2022deepphase,li2022semantic,martinez2017human,guo2023back,toyer2017human,wang2007gaussian,brand2000style,zhang2022augmented,chen2022sttg,li2022skeleton,sofianos2021space,Yan2021DMSGCNDM,Zhou2021LearningMC,Dang_2021_ICCV,Su2021MotionPV,mao2021multi,he2022reciprocal,mao2019learning,kim2022learning,cao2022qmednet,liu2022investigating,barsoum2018hp,mao2021generating,barquero2022belfusion,yuan2019diverse, bie2022hit,dilokthanakul2016deep,yan2018mt,cai2021unified,lyu2021learning,maeda2022motionaug,zhang2022pimnet,Baptiste20223dskeleton,chopin2021human,liu2022investigating,mao2021generating,punnakkal2021babel,guo2022generating,cervantes2022implicit,lehrmann2013non,delmas2022posescript,petrovich2021action,Lee2022MultiActL3,mao2022weakly}. Older popular techniques include Gaussian Processes~\cite{toyer2017human,wang2007gaussian}, Hidden Markov Models~\cite{brand2000style} leading the way to modern neural network based spatio-temporal modelling techniques~\cite{zhong2022spatial,starke2022deepphase,li2022semantic,zhang2022augmented,chen2022sttg,li2022skeleton,sofianos2021space,Yan2021DMSGCNDM,Zhou2021LearningMC,Dang_2021_ICCV,Su2021MotionPV,mao2021multi,he2022reciprocal,mao2019learning,kim2022learning,cao2022qmednet,liu2022investigating}. Recently generating diverse motion futures has gained popularity; by GANs~\cite{barsoum2018hp,mao2021generating}, and VAEs~\cite{barquero2022belfusion,yuan2019diverse, bie2022hit,dilokthanakul2016deep,yan2018mt,cai2021unified} or by treating the motion forecasting as a stochastic differential equation~\cite{lyu2021learning}. Further physically realistic human motion models~\cite{maeda2022motionaug,zhang2022pimnet}, smooth motion models~\cite{Baptiste20223dskeleton,chopin2021human,liu2022investigating,mao2021generating}
and action label and language anchored motion models have gained popularity~\cite{punnakkal2021babel,guo2022generating,cervantes2022implicit,lehrmann2013non,delmas2022posescript,petrovich2021action,Lee2022MultiActL3,mao2022weakly}.
Finally human pose forecasting in the presence of human-human interactions~\cite{katircioglu2021dyadic,vendrow2022somoformer,guo2022multi,Xu2022ActFormerAG,rahman2022pacmo,shi2022end,wang2021multi,adeli2021tripod,adeli2020socially} and human-scene interactions~\cite{fujita2023future,lee2023locomotion,Priisalu_2020_ACCV,wang2021synthesizing} have obtained increased interest but are not yet as established fields as a single human motion forecasting. The majority of modern methods utilize a high parametric NN to forecast human motion. 

\section{The problem statement}
A human pose $\bs x_i^t\in\mathcal{R}^{3,K}$ is the set of 3D coordinates (in cm) of the $K$ human joints of the individual $i$ at timestep $t$. Typical preprocessing includes normalizing the limb lengths to fit a  standardized skeleton and removing any motion of the skeleton by setting the root joint (center of the hip joints) to zero. It is also popular to represent the pose in angles formed between two limbs connected by a joint. The angles are often converted from Euler angles to an exponential map to avoid gimbal lock, but all evaluation is performed in Euler angles. A pose consists of $L$ angles $\bs z_i^t\in\mathcal{R}^{L,3}$.   
 
The standard human pose prediction problem is formulated as follows. Given a window of $M$ poses $(\bs x_i^{t-M+1},\hdots, \bs x_i^{t})$ with $M\approx 0.4s$ are observable for the prediction model to predict the $(\hat{\bs x}_i^{t+1},\hdots ,\hat{\bs x}_i^{t+N})$. The model is typically not allowed to carry memory from one prediction to another, as the sequences are randomized during the training of memory units. A model's success is typically measured by the mean per joint error
\begin{equation}
    \text{MPJE}(\hat{\bs x}_i^{t+1,}\hdots, \hat{\bs x}_i^{t+N}, \bs x_i^{t+1},\hdots, \bs x_i^{t+N})=\frac{1}{N}\frac{1}{K} \sum_{j=t+1}^{t+N}\sum_{k=1}^K\left\| \hat{\bs x}_{i,k}^{j}-\bs x_{i,k}^{j}\right\|,\label{mpje}
\end{equation}
where $\bs x_{i, k}^t\in\mathcal{R}^{3}$ are the coordinates of the $k$-th joint in the pose $\bs x_{i}^t$. For Euler angles, the Mean Euler Angle(MEA) Error is given by
\begin{equation}
    \text{MEA}(\hat{\bs z}_i^{t+1},\hdots, \hat{\bs z}_i^{t+N}, \bs z_i^{t+1},\hdots, \bs z_i^{t+N})=\frac{1}{N}\frac{1}{L} \sum_{j=t+1}^{t+N}\sum_{l=1}^L\left\| \hat{\bs z}_{i,l}^{j}-\bs z_{i,l}^{j}\right\|.\label{mea}
\end{equation}

Typical MPJE or MAE as an average over all individuals $i=(1\hdots I)$ is used as the objective function to train a NN,
\begin{equation}
    J_{MEA}=\frac{1}{I}\frac{1}{T-N} \sum_{i=1}^{I}\sum_{t=1}^{T-N}\text{MEA}(\hat{\bs z}_i^{t+1},\hdots, \hat{\bs z}_i^{t+N}, \bs z_i^{t+1},\hdots, \bs z_i^{t+N}).\label{mea_loss}
\end{equation}
When the model observes a very short time window $M$ then the optimal motion prediction according to the loss is the motion of an average pedestrian. We therefore propose to reformulate the problem by evaluating the model's ability to adapt to a single individual. Given poses $(\bs x_i^{1},\hdots, \bs x_i^{t})$ from the first timestep until the current timestep $t$ the model's ability to predict $(\hat{\bs x}_i^{t+1},\hdots, \hat{\bs x}_i^{t+N})$ is evaluated.

\section{Personalization of Neural Predictions}
Let $g$ be a model that estimates the motion of the $k$-th joint position in dimension $d$ of an average human $\hat{x}_{k,d}^{t}$ given $(\bs x_i^{t-M},\hdots, \bs x_i^{t-1})$. 
The model $g$ estimates a trend in the time series $(\bs x_i^{1}\hdots \bs x_i^{T})$. By removing the trend from all dimensions we obtain an estimated zero mean process $(y_i^{1}\hdots  y_i^{T})$, that we fit a time-varying AR model to, 
\begin{equation}
    y_{i,k,d}^{t}=-\bs\alpha_t^T \bs\phi_t+ \epsilon_t ,\label{ar}
\end{equation}
where $\bs\phi_t=(y^{t-1}_{i,k,d},\hdots, y^{t-P}_{i,k,d})$, and $\bs\alpha_t=(\alpha_{1,t}, \hdots, \alpha_{P,t})$ are time dependent model parameters and $\epsilon_t\sim\mathcal{N}(0,\sigma)$ is the innovation (white noise). The parameters $\bs \alpha_t$ are found by the prediction error method (PEM)~\cite{ljung99system}, independently for each individual, joint and dimension by optimizing 
\begin{equation}
    L_{i,k,d}=\sum_{j=1}^t\gamma^{j-t}\|y^j_{i,k,d}- \hat{y}^{j}_{i,k,d}\|^2,
\end{equation}
where $0<\gamma<1$ is the forgetting factor and the estimate $\hat{y}^{t}_{i,k,d}$ is given by linear regression $\hat{y}^{t}_{i,k,d}=\bs\alpha_t^T\bs \phi_t$ over previous samples. This weighted linear regression has the well-known closed-form solution,
\begin{equation}
    \bs \alpha_t=\left(\sum_{j=1}^t\gamma^{t-j}\bs\phi_j\bs\phi_j^T\right)^{-1} \sum_{j=1}^t\gamma^{t-j}y^{j}_{i,k,d}=\bs X_t\bs Y_t,\label{regr}
\end{equation}
where $\bs X_t^{-1}=\sum_{j=1}^t\gamma^{t-j}\bs\phi_j\bs\phi_j^t$, and $\bs Y_{t}=\sum_{j=1}^t\gamma^{t-j}y^{j}_{i,k,d}$.
To obtain an online weighted regression update we note that,
\begin{align}
    \bs X_t^{-1}&=\gamma\bs X_{t-1}^{-1}+\bs\phi_t\bs\phi_t^T\label{eqx}\\
       \bs Y_t&=\gamma\bs Y_{t-1}+\bs\phi_t y^{t}_{i,k,d}.\label{eqy}
\end{align}
Utilizing \eqref{eqx} and \eqref{eqy} in \eqref{regr} the online update for $\alpha_t$ is given as
\begin{equation}
    \bs \alpha_t=\bs \alpha_{t-1}+\bs X_t\bs \phi_t\left(y^{t}_{i,k,d}-\bs \phi_t^T\bs\phi_t\right).\label{regr}
\end{equation}
An efficient calculation of $\bs X_t^{-1}$ can be obtained from \eqref{eqx} using the matrix inversion lemma. The updates utilizing only matrix multiplications can be summarized as follows,
\begin{align}
    \bs X_t&=\frac{1}{\gamma}\bs X_{t-1}\left(I-\frac{\bs\phi_t\bs\phi_t^T\bs X_{t-1}}{\bs\phi_t^T\bs X_{t-1}\bs\phi_t+ \gamma}\right).
\end{align}

\section{Experiments}
The Human3.6M dataset is a dataset of 11 actors performing 17 actions. It contains in total 3.6 million frames. Following the previously used paradigm, we use subjects 1,6,7, and 9 for training 11 for validation, and 5 for testing. From the 32 joints in the dataset constant valued and repeated joint position dimensions are removed resulting in 48 dimensions in Euler angle representation and 66 dimensions for joint position representation. In the experiments, we follow the previous problem statement with 40ms input for and 1s of prediction unless otherwise stated. We report forecasting error, that is MPJE or MAE.  

\begin{figure}
\includegraphics[width=0.44\textwidth]{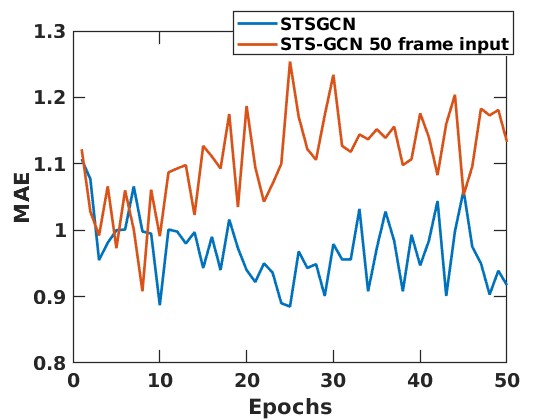}\centering
\includegraphics[width=0.44\textwidth]{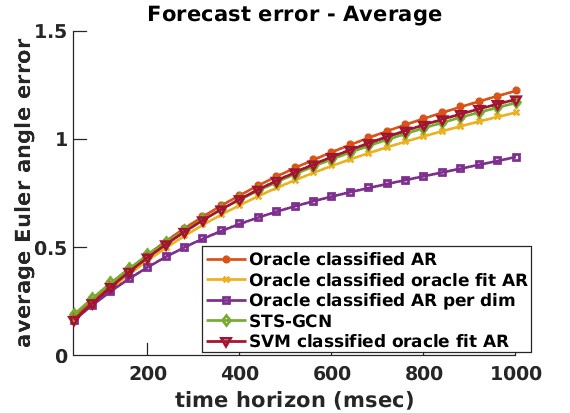}\centering
\caption{\emph{Left:} Validation errors of STS-GCN (in blue) during training show no increase in performance with an increased observation window (in orange). \emph{Right:} AR models are capable of capturing the motion information in the data and are able to outperform (see per dimension classified results in purple) STS-GCN (in green, below the SVM classified results in red and above oracle fit and classified AR) on the Human3.6M benchmark.} \label{personalization_oracle1}
\end{figure}
We study the influential Space-Time-Separable Graph Convolutional Network for Pose Forecasting (STS-GCN)~\cite{sofianos2021space} that showed 30\% improvement over previous work with fewer parameters on the popular Human3.6M~\cite{ionescu2013human3} dataset. STS-GCN~\cite{sofianos2021space} use a Graph Convolutional Network(GCN) to learn the temporal and spatial features in the 3D poses with an explicit separation of the spatial and temporal relations into separate adjacency matrices modeled by per channel and across channel convolutions. Predictions are extracted from the GCN features with a Temporal Convolution Network (TCN). A TCN performs causal convolutions over the time domain.  

The STS-GCN fails to learn the personalized behaviors when given an observation window that is twice as long as the prediction window as seen in \Figure{personalization_oracle1} \emph{left}. It is expected that STS-GCN should be able to learn any periodic behaviors when the observation window is larger than the prediction time horizon, but no significant improvement can be observed. This illustrates that NN training can be temperamental and a new architecture search is necessary for a change in time windows in the problem statement. 

\begin{figure}
\includegraphics[width=0.44\textwidth]{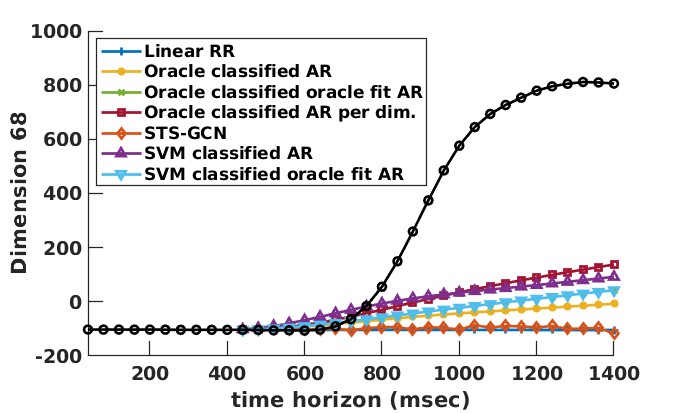}\centering
\includegraphics[width=0.44\textwidth]{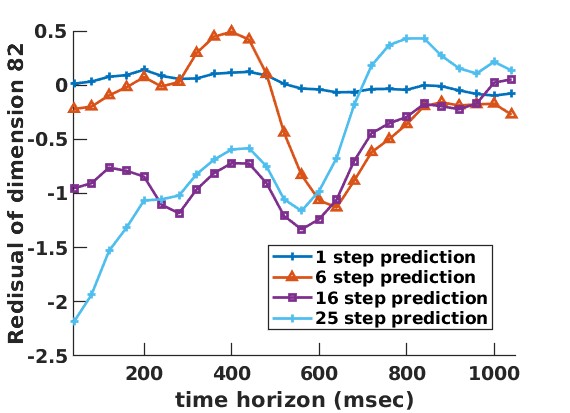}\centering
\caption{\emph{Left:}The \emph{STS-GCN} (in orange at the bottom) fails at times to foresee sudden future motion (ground truth in black), while the AR-based models predict an increase in dimension 68. \emph{Right:} Residuals of \emph{STS-GCN} increase in amplitude with increasing prediction horizons. One-step residuals(in dark blue) are almost flat, while the residuals of the 25-step prediction in light blue oscillate.} \label{personalization_oracle}
\end{figure}

To see the ability of AR to model human dynamics a set of AR models $\bs f_i$ (consisting of one AR model per dimension) is fitted to each individual in the training data (model order chosen by Bayesian Information Content). A classifier (oracle or a linear support vector machine) is trained to select the $i$-th individual's set $\bs f_i$ among the ones trained on the training set that results in the lowest prediction error on the validation set. The STS-GCN is compared with the following models,   
\begin{itemize}
    \item \emph{Oracle classified AR} - an oracle classifier selects the $\bs f_i$ that has the lowest prediction error.
     \item \emph{Oracle classified oracle fitted AR} - an oracle classifier selects the lag structure of $\bs f_i$ that has the lowest prediction error. A new set of AR model $\hat{\bs f_i}$ that has the same lag per dimension as $\bs f_i$ is fitted to the test data.
      \item \emph{Oracle classified AR per dimension} - an oracle classifier selects for each dimension $d$ the $\bs f_{i,d}$ that has the lowest prediction error.
       \item \emph{SVM classified oracle fitted AR} - a support vector machine selects the $\bs f_i$ that has the lowest prediction error.
\end{itemize}
Utilizing \emph{oracle fitting} and \emph{oracle classification} shows the lower error bounds obtainable. In \Figure{personalization_oracle1} \emph{left} it can be seen that STS-GCN has a performance between that of \emph{oracle-classified AR models}, \emph{SVM-classified AR models}, and \emph{oracle-classified and oracle-fitted AR models}. The fact that refitting the AR model on the test data does not improve the performance much suggests that the AR model structure is more important than the exact parameter values. Making small errors on lag estimation does not bring performance down much as seen by comparing the \emph{SVM-classified oracle fit AR models} performance with that of \emph{Oracle classified Oracle fit AR models} in \Figure{personalization_oracle1} \emph{left}. The AR model structures appear not to be transferable across skeletons but across joints, as selecting the AR model per dimension shows a great improvement over selecting AR models per person or the \emph{STS-GCN}, compare \emph{Oracle classified AR per dimension} with  \emph{Oracle classified AR} and \emph{STS-GCN} \Figure{personalization_oracle1} \emph{left}. This illustrates the need for personalizing the motion model to new individuals as new relations between joints need to be learned when transferring between people. Sometimes \emph{STS-GCN} fails to foresee sudden motion, when AR-based methods expect a change as seen in \Figure{personalization_oracle} \emph{Left}. 

Human motion is in general non-stationary, but the 1-step prediction residuals in \Figure{personalization_oracle} \emph{Right} of \emph{STS-GCN }look well behaved. Therefore having observed that AR models are capable of modeling human behavior with a performance that is comparable to NNs we utilize \emph{STS-GCN} to remove the trend in the data and fit a time-varying AR process to the residuals. This will be referred to as the \emph{recursive AR model}.
  
\begin{figure}
\includegraphics[width=0.44\textwidth]{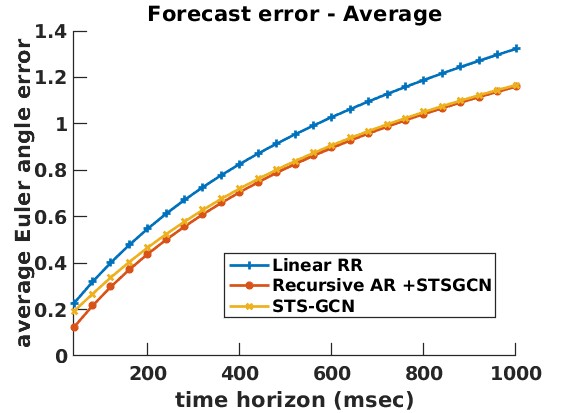}\centering
\includegraphics[width=0.44\textwidth]{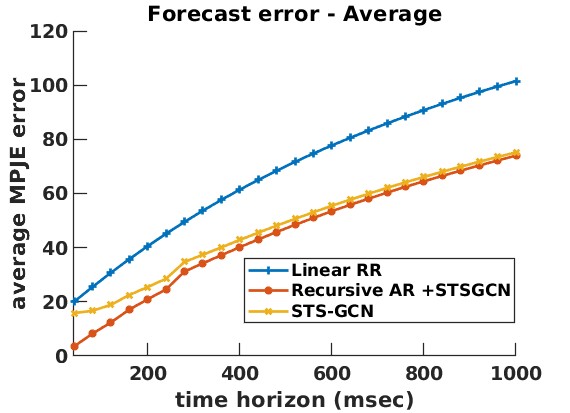}\centering
\caption{\emph{Left:} An online-estimated AR(1) model (in yellow) corrects the early errors of \emph{STS-GCN} (in orange) on angular data. \emph{Linear Ridge Regression} forecasting errors (in blue) are for reference. \emph{Right:}An online-estimated AR(1) model corrects the early errors of \emph{STS-GCN} on positional data.} \label{personalization2}
\end{figure}

In \Figure{personalization2} it can be seen that the \emph{recursive AR model} improves the \emph{STS-GCN } in early steps where the AR model has the most effect. It can also be seen that \emph{STS-GCN} does not improve particularly much over \emph{linear ridge regression}. Other AR structures showed no improvement over the lag 1 time-varying AR models on the \emph{STS-GCN} residuals.
\subsection{Discussion and Future Work}
The proposed method of utilizing AR models to personalize human pose forecasting provides a low-dimensional model that can be trained online. This is however only a first step in utilizing time series analysis to personalize human pose forecasting because the AR model's exponentially decreasing prediction power doesn't allow long-range predictions to be adapted. To adapt long-term predictions a longer time horizon must be observed and a model-specific adaptation must be performed such as meta-learning of personalized key-pose predictions on key-pose-based human motion forecasting methods~\cite{diller2022charposes,kiciroglu2022keyposes}. It should be noted that AR models obtain comparable performance to neural networks with much fewer parameters (~50-500 vs 14k and up). In the future AR models could be utilized for fast and efficient human motion forecasting by using meta-learning to estimate the AR model structure. AR models being low parametric have low hardware requirements and could therefore be beneficial on mobile devices. 
\section{Conclusion}
Existing methods for human motion forecasting utilize neural networks that are trained to optimize to guess the motion of an average human due to the short observation window. In practice, HCI systems interact with a single individual for a longer time than is reflected in existing human motion forecasting benchmarks and model prediction  should adapt to the individual over the course of the interaction. Human motion is highly individual so optimally a model should adapt forecasts at test time to new individuals. Updating existing neural models to new individuals during test time is not trivial. We provide a lightweight and generalizable solution by fitting a low parametric model to the neural model's residuals. Recursive AR models can be used to personalize existing neural networks based human motion models. AR models are low parametric and can be fitted online avoiding possible issues in online neural network updates. We show that AR models have the capacity to estimate human motion on par with neural networks with much fewer parameters. Combining time series analysis and neural models could lead to new research directions in the field in particular for on-device solutions.
 
\printbibliography

@article{ionescu2013human3,
  title={Human3. 6m: Large scale datasets and predictive methods for 3d human sensing in natural environments},
  author={Ionescu, Catalin and Papava, Dragos and Olaru, Vlad and Sminchisescu, Cristian},
  journal={IEEE Transactions on Pattern Analysis and Machine Intelligence},
  volume={36},
  number={7},
  pages={1325--1339},
  year={2013},
  publisher={IEEE}
}

@article{sigal2010humaneva,
  title={Humaneva: Synchronized video and motion capture dataset and baseline algorithm for evaluation of articulated human motion},
  author={Sigal, Leonid and Balan, Alexandru O and Black, Michael J},
  journal={International Journal of Computer Vision},
  volume={87},
  number={1},
  pages={4--27},
  year={2010},
  publisher={Springer}
}

@conference{AMASS:ICCV:2019,
  title = {{AMASS}: Archive of Motion Capture as Surface Shapes},
  author = {Mahmood, Naureen and Ghorbani, Nima and Troje, Nikolaus F. and Pons-Moll, Gerard and Black, Michael J.},
  booktitle = {ICCV},
  pages = {5442--5451},
  month = oct,
  year = {2019},
  month_numeric = {10}
}

@online{CMUMoCaP,
  author = {CMU Graphics Lab},
  title = {“CMU Graphics Lab Motion Capture Database.”},
  url = {http://mocap.cs.cmu.edu},
  urldate = {2023-01-16}
}

@book{box2015time,
  title={Time series analysis: forecasting and control},
  author={Box, George EP and Jenkins, Gwilym M and Reinsel, Gregory C and Ljung, Greta M},
  year={2015},
  publisher={John Wiley \& Sons}
}

@article{tallamraju2020aircaprl,
  title={AirCapRL: autonomous aerial human motion capture using deep reinforcement learning},
  author={Tallamraju, Rahul and Saini, Nitin and Bonetto, Elia and Pabst, Michael and Liu, Yu Tang and Black, Michael J and Ahmad, Aamir},
  journal={IEEE Robotics and Automation Letters},
  volume={5},
  number={4},
  pages={6678--6685},
  year={2020},
  publisher={IEEE}
}

@article{palermo2022raw,
  title={From raw measurements to human pose-a dataset with low-cost and high-end inertial-magnetic sensor data},
  author={Palermo, Manuel and Cerqueira, Sara M and Andr{\'e}, Jo{\~a}o and Pereira, Ant{\'o}nio and Santos, Cristina P},
  journal={Scientific Data},
  volume={9},
  number={1},
  pages={1--9},
  year={2022},
  publisher={Nature Publishing Group}
}

@inproceedings{von2018recovering,
  title={Recovering accurate 3d human pose in the wild using imus and a moving camera},
  author={Von Marcard, Timo and Henschel, Roberto and Black, Michael J and Rosenhahn, Bodo and Pons-Moll, Gerard},
  booktitle={Proceedings of the ECCV },
  pages={601--617},
  year={2018}
}

@inproceedings{shukla2022vl4pose,
  title={VL4Pose: Active Learning Through Out-Of-Distribution Detection For Pose Estimation},
  author={Shukla, Megh and Roy, Roshan and Singh, Pankaj and Ahmed, Shuaib and Alahi, Alexandre},
  booktitle={Proceedings of the 33rd BMVC},
  number={CONF},
  year={2022},
  organization={BMVA Press}
}

@inproceedings{kiciroglu2022keyposes,
  author = {Kiciroglu, Sena and Wang, Wei and Salzmann, Mathieu and Fua, Pascal},
  booktitle = {3DV},
  title = {Long Term Motion Prediction Using Keyposes},
  year = {2022}
}

@inproceedings{ma2022progressively,
  title={Progressively Generating Better Initial Guesses Towards Next Stages for High-Quality Human Motion Prediction},
  author={Ma, Tiezheng and Nie, Yongwei and Long, Chengjiang and Zhang, Qing and Li, Guiqing},
  booktitle={Proceedings of the IEEE/CVF Conference on CVPR},
  pages={6437--6446},
  year={2022}
}

@article{diller2022charposes,
    title={Forecasting Characteristic 3D Poses of Human Actions},
    author={Diller, Christian and Funkhouser, Thomas and Dai, Angela},
    booktitle={Proc. CVPR, IEEE},
    year={2022}
}

@article{zang2022few,
  title={Few-shot human motion prediction using deformable spatio-temporal CNN with parameter generation},
  author={Zang, Chuanqi and Li, Menghao and Pei, Mingtao},
  journal={Neurocomputing},
  volume={513},
  pages={46--58},
  year={2022},
  publisher={Elsevier}
}

@inproceedings{xu2021probabilistic,
  title={Probabilistic human motion prediction via A bayesian neural network},
  author={Xu, Jie and Chen, Xingyu and Lan, Xuguang and Zheng, Nanning},
  booktitle={2021 IEEE ICRA},
  pages={3190--3196},
  year={2021},
  organization={IEEE}
}

@article{ding2021uncertainty,
  title={Uncertainty-aware Human Motion Prediction},
  author={Ding, Pengxiang and Yin, Jianqin},
  journal={arXiv preprint arXiv:2107.03575},
  year={2021}
}

@inproceedings{saadatnejad2022generic,
  title={A generic diffusion-based approach for 3D human pose prediction in the wild},
  author={Saadatnejad, Saeed and Rasekh, Ali and Mofayezi, Mohammadreza and Medghalchi, Yasamin and Rajabzadeh, Sara and Mordan, Taylor and Alahi, Alexandre},
  booktitle={NeurIPS 2022 Workshop on Score-Based Methods}
}

@article{drumond2022few,
  title={Few-shot human motion prediction for heterogeneous sensors},
  author={Drumond, Rafael Rego and Brinkmeyer, Lukas and Schmidt-Thieme, Lars},
  journal={arXiv preprint arXiv:2212.11771},
  year={2022}
}

@article{kieu2022locally,
  title={Locally weighted PCA regression to recover missing markers in human motion data},
  author={Kieu, Hai Dang and Yu, Hongchuan and Li, Zhuorong and Zhang, Jian Jun},
  journal={Plos one},
  volume={17},
  number={8},
  pages={e0272407},
  year={2022},
  publisher={Public Library of Science San Francisco, CA USA}
}

@inproceedings{Gui2018FewShotHM,
  title={Few-Shot Human Motion Prediction via Meta-learning},
  author={Liangyan Gui and Yu-Xiong Wang and Deva Ramanan and Jos{\'e} M. F. Moura},
  booktitle={ECCV},
  year={2018}
}

@article{balazia2018gait,
  title={Gait recognition from motion capture data},
  author={Balazia, Michal and Sojka, Petr},
  journal={ACM Transactions on Multimedia Computing, Communications, and Applications (TOMM)},
  volume={14},
  number={1s},
  pages={1--18},
  year={2018},
  publisher={ACM New York, NY, USA}
}

@inproceedings{zhu2022personalized,
  title={Personalized individual trajectory prediction via meta-learning},
  author={Zhu, He and Zhang, Liyu and Fan, Zipei},
  booktitle={Proceedings of the 30th International Conference on Advances in Geographic Information Systems},
  pages={1--2},
  year={2022}
}

@article{wan2018survey,
  title={A survey on gait recognition},
  author={Wan, Changsheng and Wang, Li and Phoha, Vir V},
  journal={ACM Computing Surveys (CSUR)},
  volume={51},
  number={5},
  pages={1--35},
  year={2018},
  publisher={ACM New York, NY, USA}
}

@article{singh2018vision,
  title={Vision-based gait recognition: A survey},
  author={Singh, Jasvinder Pal and Jain, Sanjeev and Arora, Sakshi and Singh, Uday Pratap},
  journal={Ieee Access},
  volume={6},
  pages={70497--70527},
  year={2018},
  publisher={IEEE}
}

@article{nikdel2022dmmgan,
  title={DMMGAN: Diverse Multi Motion Prediction of 3D Human Joints using Attention-Based Generative Adverserial Network},
  author={Nikdel, Payam and Mahdavian, Mohammad and Chen, Mo},
  journal={arXiv preprint arXiv:2209.09124},
  year={2022}
}

@inproceedings{adeli2021tripod,
  title={Tripod: Human trajectory and pose dynamics forecasting in the wild},
  author={Adeli, Vida and Ehsanpour, Mahsa and Reid, Ian and Niebles, Juan Carlos and Savarese, Silvio and Adeli, Ehsan and Rezatofighi, Hamid},
  booktitle={Proceedings of the IEEE/CVF ICCV},
  pages={13390--13400},
  year={2021}
}

@article{adeli2020socially,
  title={Socially and contextually aware human motion and pose forecasting},
  author={Adeli, Vida and Adeli, Ehsan and Reid, Ian and Niebles, Juan Carlos and Rezatofighi, Hamid},
  journal={IEEE Robotics and Automation Letters},
  volume={5},
  number={4},
  pages={6033--6040},
  year={2020},
  publisher={IEEE}
}

@inproceedings{sampieri2022pose,
  title={Pose Forecasting in Industrial Human-Robot Collaboration},
  author={Sampieri, Alessio and di Melendugno, Guido Maria D’Amely and Avogaro, Andrea and Cunico, Federico and Setti, Francesco and Skenderi, Geri and Cristani, Marco and Galasso, Fabio},
  booktitle={ECCV},
  pages={51--69},
  year={2022},
  organization={Springer}
}

@inproceedings{toyer2017human,
  title={Human pose forecasting via deep markov models},
  author={Toyer, Sam and Cherian, Anoop and Han, Tengda and Gould, Stephen},
  booktitle={2017 International Conference on Digital Image Computing: Techniques and Applications (DICTA)},
  pages={1--8},
  year={2017},
  organization={IEEE}
}

@inproceedings{brand2000style,
  title={Style machines},
  author={Brand, Matthew and Hertzmann, Aaron},
  booktitle={Proceedings of the 27th annual conference on Computer graphics and interactive techniques},
  pages={183--192},
  year={2000}
}

@article{wang2007gaussian,
  title={Gaussian process dynamical models for human motion},
  author={Wang, Jack M and Fleet, David J and Hertzmann, Aaron},
  journal={IEEE transactions on pattern analysis and machine intelligence},
  volume={30},
  number={2},
  pages={283--298},
  year={2007},
  publisher={IEEE}
}

@article{barquero2022belfusion,
  title={BeLFusion: Latent Diffusion for Behavior-Driven Human Motion Prediction},
  author={Barquero, German and Escalera, Sergio and Palmero, Cristina},
  journal={arXiv preprint arXiv:2211.14304},
  year={2022}
}

@inproceedings{yuan2019diverse,
  title={Diverse Trajectory Forecasting with Determinantal Point Processes},
  author={Yuan, Ye and Kitani, Kris M},
  booktitle={ICLR},
  year={2019}
}

@article{bie2022hit,
  title={HiT-DVAE: Human Motion Generation via Hierarchical Transformer Dynamical VAE},
  author={Bie, Xiaoyu and Guo, Wen and Leglaive, Simon and Girin, Lauren and Moreno-Noguer, Francesc and Alameda-Pineda, Xavier},
  journal={arXiv preprint arXiv:2204.01565},
  year={2022}
}

@article{dilokthanakul2016deep,
  title={Deep unsupervised clustering with gaussian mixture variational autoencoders},
  author={Dilokthanakul, Nat and Mediano, Pedro AM and Garnelo, Marta and Lee, Matthew CH and Salimbeni, Hugh and Arulkumaran, Kai and Shanahan, Murray},
  journal={arXiv preprint arXiv:1611.02648},
  year={2016}
}

@inproceedings{yan2018mt,
  title={Mt-vae: Learning motion transformations to generate multimodal human dynamics},
  author={Yan, Xinchen and Rastogi, Akash and Villegas, Ruben and Sunkavalli, Kalyan and Shechtman, Eli and Hadap, Sunil and Yumer, Ersin and Lee, Honglak},
  booktitle={Proceedings of the ECCV },
  pages={265--281},
  year={2018}
}

@inproceedings{barsoum2018hp,
  title={Hp-gan: Probabilistic 3d human motion prediction via gan},
  author={Barsoum, Emad and Kender, John and Liu, Zicheng},
  booktitle={Proceedings of the IEEE conference on CVPR workshops},
  pages={1418--1427},
  year={2018}
}

@inproceedings{mao2021generating,
  title={Generating smooth pose sequences for diverse human motion prediction},
  author={Mao, Wei and Liu, Miaomiao and Salzmann, Mathieu},
  booktitle={Proceedings of the IEEE/CVF ICCV},
  pages={13309--13318},
  year={2021}
}

@inproceedings{lyu2021learning,
  title={Learning human motion prediction via stochastic differential equations},
  author={Lyu, Kedi and Liu, Zhenguang and Wu, Shuang and Chen, Haipeng and Zhang, Xuhong and Yin, Yuyu},
  booktitle={Proceedings of the 29th ACM International Conference on Multimedia},
  pages={4976--4984},
  year={2021}
}

@inproceedings{cai2021unified,
  title={A unified 3d human motion synthesis model via conditional variational auto-encoder},
  author={Cai, Yujun and Wang, Yiwei and Zhu, Yiheng and Cham, Tat-Jen and Cai, Jianfei and Yuan, Junsong and Liu, Jun and Zheng, Chuanxia and Yan, Sijie and Ding, Henghui and others},
  booktitle={Proceedings of the IEEE/CVF ICCV},
  pages={11645--11655},
  year={2021}
}

@article{zhong2022spatial,
  title={Spatial--temporal modeling for prediction of stylized human motion},
  author={Zhong, Chongyang and Hu, Lei and Xia, Shihong},
  journal={Neurocomputing},
  volume={511},
  pages={34--42},
  year={2022},
  publisher={Elsevier}
}

@article{starke2022deepphase,
  title={Deepphase: Periodic autoencoders for learning motion phase manifolds},
  author={Starke, Sebastian and Mason, Ian and Komura, Taku},
  journal={ACM TOG},
  volume={41},
  number={4},
  pages={1--13},
  year={2022},
  publisher={ACM New York, NY, USA}
}

@article{li2022semantic,
  title={Semantic Correlation Attention-Based Multiorder Multiscale Feature Fusion Network for Human Motion Prediction},
  author={Li, Qin and Wang, Yong and Lv, Fanbing},
  journal={IEEE Transactions on Cybernetics},
  year={2022},
  publisher={IEEE}
}

@inproceedings{zhang2022augmented,
  title={Augmented Graph Attention with Temporal Gradation and Reorganization for Human Motion Prediction},
  author={Zhang, Shaobo and Liu, Sheng and Gao, Fei and Chen, Shengyong},
  booktitle={International Conference on Intelligent Robotics and Applications},
  pages={294--306},
  year={2022},
  organization={Springer}
}

@article{chen2022sttg,
  title={STTG-net: a Spatio-temporal network for human motion prediction based on transformer and graph convolution network},
  author={Chen, Lujing and Liu, Rui and Yang, Xin and Zhou, Dongsheng and Zhang, Qiang and Wei, Xiaopeng},
  journal={Visual Computing for Industry, Biomedicine, and Art},
  volume={5},
  number={1},
  pages={1--15},
  year={2022},
  publisher={Springer}
}

@inproceedings{li2022skeleton,
  title={Skeleton-Parted Graph Scattering Networks for 3D Human Motion Prediction},
  author={Li, Maosen and Chen, Siheng and Zhang, Zijing and Xie, Lingxi and Tian, Qi and Zhang, Ya},
  booktitle={ECCV},
  pages={18--36},
  year={2022},
  organization={Springer}
}

@inproceedings{sofianos2021space,
  title={Space-time-separable graph convolutional network for pose forecasting},
  author={Sofianos, Theodoros and Sampieri, Alessio and Franco, Luca and Galasso, Fabio},
  booktitle={Proceedings of the IEEE/CVF ICCV},
  pages={11209--11218},
  year={2021}
}

@article{Yan2021DMSGCNDM,
  title={DMS-GCN: Dynamic Mutiscale Spatiotemporal Graph Convolutional Networks for Human Motion Prediction},
  author={Zigeng Yan and Dihua Zhai and Yuanqing Xia},
  journal={arXiv preprint arXiv:2112.10365},
  year={2021}
}

@article{Zhou2021LearningMC,
  title={Learning Multiscale Correlations for Human Motion Prediction},
  author={Honghong Zhou and Caili Guo and Hao Zhang and Yanjun Wang},
  journal={2021 IEEE International Conference on Development and Learning (ICDL)},
  year={2021},
  pages={1-7}
}

@InProceedings{Dang_2021_ICCV,
    author    = {Dang, Lingwei and Nie, Yongwei and Long, Chengjiang and Zhang, Qing and Li, Guiqing},
    title     = {MSR-GCN: Multi-Scale Residual Graph Convolution Networks for Human Motion Prediction},
    booktitle = {Proceedings of the IEEE/CVF ICCV (ICCV)},
    month     = {October},
    year      = {2021},
    pages     = {11467-11476}
}

@article{Su2021MotionPV,
  title={Motion Prediction via Joint Dependency Modeling in Phase Space},
  author={Pengxiang Su and Zhenguang Liu and Shuang Wu and Lei Zhu and Yifang Yin and Xuanjing Shen},
  journal={Proceedings of the 29th ACM International Conference on Multimedia},
  year={2021}
}

@article{mao2021multi,
  title={Multi-level motion attention for human motion prediction},
  author={Mao, Wei and Liu, Miaomiao and Salzmann, Mathieu and Li, Hongdong},
  journal={International Journal of Computer Vision},
  volume={129},
  number={9},
  pages={2513--2535},
  year={2021},
  publisher={Springer}
}

@inproceedings{he2022reciprocal,
  title={Reciprocal collaboration network for 3D skeleton-based human motion prediction},
  author={He, Zhiquan and Zhang, Lujun and Cao, Wenming},
  booktitle={Third International Conference on Computer Science and Communication Technology (ICCSCT 2022)},
  volume={12506},
  pages={935--942},
  year={2022},
  organization={SPIE}
}

@inproceedings{mao2019learning,
  title={Learning trajectory dependencies for human motion prediction},
  author={Mao, Wei and Liu, Miaomiao and Salzmann, Mathieu and Li, Hongdong},
  booktitle={Proceedings of the IEEE/CVF ICCV},
  pages={9489--9497},
  year={2019}
}

@article{kim2022learning,
  title={Learning spectral transform for 3D human motion prediction},
  author={Kim, Boeun and Choi, Jin Young},
  journal={Computer Vision and Image Understanding},
  volume={223},
  pages={103548},
  year={2022},
  publisher={Elsevier}
}

@article{cao2022qmednet,
  title={QMEDNet: A quaternion-based multi-order differential encoder--decoder model for 3D human motion prediction},
  author={Cao, Wenming and Li, Shuangshuang and Zhong, Jianqi},
  journal={Neural Networks},
  volume={154},
  pages={141--151},
  year={2022},
  publisher={Elsevier}
}

@inproceedings{martinez2017human,
  title={On human motion prediction using recurrent neural networks},
  author={Martinez, Julieta and Black, Michael J and Romero, Javier},
  booktitle={Proceedings of the IEEE conference on CVPR},
  pages={2891--2900},
  year={2017}
}

@inproceedings{guo2023back,
  title={Back to mlp: A simple baseline for human motion prediction},
  author={Guo, Wen and Du, Yuming and Shen, Xi and Lepetit, Vincent and Alameda-Pineda, Xavier and Moreno-Noguer, Francesc},
  booktitle={Proceedings of the IEEE/CVF Winter Conference on Applications of Computer Vision},
  pages={4809--4819},
  year={2023}
}

@article{liu2022investigating,
  title={Investigating pose representations and motion contexts modeling for 3D motion prediction},
  author={Liu, Zhenguang and Wu, Shuang and Jin, Shuyuan and Ji, Shouling and Liu, Qi and Lu, Shijian and Cheng, Li},
  journal={IEEE Transactions on Pattern Analysis and Machine Intelligence},
  volume={45},
  number={1},
  pages={681--697},
  year={2022},
  publisher={IEEE}
}

@inproceedings{maeda2022motionaug,
  title={MotionAug: Augmentation with Physical Correction for Human Motion Prediction},
  author={Maeda, Takahiro and Ukita, Norimichi},
  booktitle={Proceedings of the IEEE/CVF Conference on CVPR},
  pages={6427--6436},
  year={2022}
}

@article{zhang2022pimnet,
  title={PIMNet: Physics-Infused Neural Network for Human Motion Prediction},
  author={Zhang, Zhibo and Zhu, Yanjun and Rai, Rahul and Doermann, David},
  journal={IEEE Robotics and Automation Letters},
  volume={7},
  number={4},
  pages={8949--8955},
  year={2022},
  publisher={IEEE}
}

@article{Baptiste20223dskeleton,
author = {Chopin, Baptiste and Otberdout, Naima and Daoudi, Mohamed and Bartolo, Angela},
year = {2022},
month = {10},
pages = {1-1},
title = {3D Skeleton-based Human Motion Prediction with Manifold-Aware GAN},
volume = {PP},
journal = {IEEE Transactions on Biometrics, Behavior, and Identity Science},
}

@inproceedings{chopin2021human,
  title={Human Motion Prediction Using Manifold-Aware Wasserstein GAN},
  author={Chopin, Baptiste and Otberdout, Naima and Daoudi, Mohamed and Bartolo, Angela},
  booktitle={2021 16th IEEE International Conference on Automatic Face and Gesture Recognition (FG 2021)},
  pages={1--8},
  year={2021},
  organization={IEEE}
}

@inproceedings{punnakkal2021babel,
  title={BABEL: Bodies, action and behavior with english labels},
  author={Punnakkal, Abhinanda R and Chandrasekaran, Arjun and Athanasiou, Nikos and Quiros-Ramirez, Alejandra and Black, Michael J},
  booktitle={Proceedings of the IEEE/CVF Conference on CVPR},
  pages={722--731},
  year={2021}
}

@inproceedings{guo2022generating,
  title={Generating Diverse and Natural 3D Human Motions From Text},
  author={Guo, Chuan and Zou, Shihao and Zuo, Xinxin and Wang, Sen and Ji, Wei and Li, Xingyu and Cheng, Li},
  booktitle={Proceedings of the IEEE/CVF Conference on CVPR},
  pages={5152--5161},
  year={2022}
}

@inproceedings{cervantes2022implicit,
  title={Implicit neural representations for variable length human motion generation},
  author={Cervantes, Pablo and Sekikawa, Yusuke and Sato, Ikuro and Shinoda, Koichi},
  booktitle={ECCV},
  pages={356--372},
  year={2022},
  organization={Springer}
}

@inproceedings{lehrmann2013non,
  title={A non-parametric bayesian network prior of human pose},
  author={Lehrmann, Andreas M and Gehler, Peter V and Nowozin, Sebastian},
  booktitle={Proceedings of the IEEE ICCV},
  pages={1281--1288},
  year={2013}
}

@inproceedings{delmas2022posescript,
  title={PoseScript: 3D human poses from natural language},
  author={Delmas, Ginger and Weinzaepfel, Philippe and Lucas, Thomas and Moreno-Noguer, Francesc and Rogez, Gr{\'e}gory},
  booktitle={ECCV},
  pages={346--362},
  year={2022},
  organization={Springer}
}

@inproceedings{petrovich2021action,
  title={Action-conditioned 3d human motion synthesis with transformer vae},
  author={Petrovich, Mathis and Black, Michael J and Varol, G{\"u}l},
  booktitle={Proceedings of the IEEE/CVF ICCV},
  pages={10985--10995},
  year={2021}
}

@article{Lee2022MultiActL3,
  title={MultiAct: Long-Term 3D Human Motion Generation from Multiple Action Labels},
  author={Tae Jun (David) Lee and Gyeongsik Moon and Kyoung Mu Lee},
  number={5}, 
  journal={Proceedings of the AAAI Conference on Artificial Intelligence},
  year={2023},
  volume={tba}
}

@inproceedings{mao2022weakly,
  title={Weakly-supervised Action Transition Learning for Stochastic Human Motion Prediction},
  author={Mao, Wei and Liu, Miaomiao and Salzmann, Mathieu},
  booktitle={Proceedings of the IEEE/CVF Conference on CVPR},
  pages={8151--8160},
  year={2022}
}

@article{lyu20223d,
  title={3D human motion prediction: A survey},
  author={Lyu, Kedi and Chen, Haipeng and Liu, Zhenguang and Zhang, Beiqi and Wang, Ruili},
  journal={Neurocomputing},
  volume={489},
  pages={345--365},
  year={2022},
  publisher={Elsevier}
}

@article{katircioglu2021dyadic,
  title={Dyadic human motion prediction},
  author={Katircioglu, Isinsu and Georgantas, Costa and Salzmann, Mathieu and Fua, Pascal},
  journal={arXiv preprint arXiv:2112.00396},
  year={2021}
}

@article{vendrow2022somoformer,
  title={SoMoFormer: Multi-Person Pose Forecasting with Transformers},
  author={Vendrow, Edward and Kumar, Satyajit and Adeli, Ehsan and Rezatofighi, Hamid},
  journal={arXiv preprint arXiv:2208.14023},
  year={2022}
}

@inproceedings{guo2022multi,
  title={Multi-Person Extreme Motion Prediction},
  author={Guo, Wen and Bie, Xiaoyu and Alameda-Pineda, Xavier and Moreno-Noguer, Francesc},
  booktitle={Proceedings of the IEEE/CVF Conference on CVPR},
  pages={13053--13064},
  year={2022}
}

@inproceedings{Xu2022ActFormerAG,
  title={ActFormer: A GAN-based Transformer towards General Action-Conditioned 3D Human Motion Generation},
  author={Liang Xu and Ziyang Song and Dongliang Wang and Jing Su and Zhicheng Fang and Chen Ding and Weihao Gan and Yichao Yan and Xin Jin and Xiaokang Yang and Wenjun Zeng and Wei Wu},
  year={2022}
}

@article{rahman2022pacmo,
  title={PaCMO: Partner Dependent Human Motion Generation in Dyadic Human Activity using Neural Operators},
  author={Rahman, Md Ashiqur and Ghosh, Jasorsi and Viswanath, Hrishikesh and Azizzadenesheli, Kamyar and Bera, Aniket},
  journal={arXiv preprint arXiv:2211.16210},
  year={2022}
}

@inproceedings{shi2022end,
  title={End-to-End Multi-Person Pose Estimation With Transformers},
  author={Shi, Dahu and Wei, Xing and Li, Liangqi and Ren, Ye and Tan, Wenming},
  booktitle={Proceedings of the IEEE/CVF Conference on CVPR},
  pages={11069--11078},
  year={2022}
}

@article{wang2021multi,
  title={Multi-Person 3D Motion Prediction with Multi-Range Transformers},
  author={Wang, Jiashun and Xu, Huazhe and Narasimhan, Medhini and Wang, Xiaolong},
  journal={Advances in Neural Information Processing Systems},
  volume={34},
  pages={6036--6049},
  year={2021}
}

@article{fujita2023future,
  title={Future Pose Prediction from 3D Human Skeleton Sequence with Surrounding Situation},
  author={Fujita, Tomohiro and Kawanishi, Yasutomo},
  journal={Sensors},
  volume={23},
  number={2},
  pages={876},
  year={2023},
  publisher={Multidisciplinary Digital Publishing Institute}
}

@article{lee2023locomotion,
  title={Locomotion-Action-Manipulation: Synthesizing Human-Scene Interactions in Complex 3D Environments},
  author={Lee, Jiye and Joo, Hanbyul},
  journal={arXiv preprint arXiv:2301.\-02667},
  year={2023}
}

@inproceedings{Priisalu_2020_ACCV,
    author    = {Priisalu, Maria and Paduraru, Ciprian and Pirinen, Aleksis and Sminchisescu, Cristian},
    title     = {Semantic Synthesis of Pedestrian Locomotion},
    booktitle = {Proceedings of the ACCV},
    month     = {November},
    year      = {2020}
}

@inproceedings{wang2021synthesizing,
  title={Synthesizing long-term 3d human motion and interaction in 3d scenes},
  author={Wang, Jiashun and Xu, Huazhe and Xu, Jingwei and Liu, Sifei and Wang, Xiaolong},
  booktitle={Proceedings of the IEEE/CVF Conference on CVPR},
  pages={9401--9411},
  year={2021}
}

@incollection{mccloskey1989catastrophic,
  title={Catastrophic interference in connectionist networks: The sequential learning problem},
  author={McCloskey, Michael and Cohen, Neal J},
  booktitle={Psychology of learning and motivation},
  volume={24},
  pages={109--165},
  year={1989},
  publisher={Elsevier}
}

@book{ljung99system,
Author = {Ljung, Lennart},
Publisher = {Prentice Hall},
Series = {Prentice-Hall information and system sciences series},
Title = {System identification : theory for the user.},
Year = {1999},
}

@article{CAO2022141,
title = {QMEDNet: A quaternion-based multi-order differential encoder–decoder model for 3D human motion prediction},
journal = {Neural Networks},
volume = {154},
pages = {141-151},
year = {2022},
author = {Wenming Cao and Shuangshuang Li and Jianqi Zhong},
}

@article{Grassia98,
  author    = {F. Sebastian Grassia},
  title     = {Practical Parameterization of Rotations Using the Exponential Map},
  journal   = {J. Graphics, GPU, {\&} Game Tools},
  volume    = {3},
  number    = {3},
  pages     = {29--48},
  year      = {1998}
}
\appendix
\section{Supplementary material}
\subsection{3D Pose Representation in Angles}
Human poses in 3D can be represented by joint positions of angles. Here a common treatment of the angular representation is given. There exist methods that utilize the quaternion representation directly~\cite{CAO2022141} or even learn their own spectral transforms~\cite{kim2022learning}.
It is a common treatment to center the skeleton around the hip joint and to rescale the joint positions to fit a uniform skeleton (with average joint lengths). As a result of these two actions, a number of the joint angles have zero value. The human skeleton is seen as a tree graph. The rotation of a limb is given by the product of the joint's parents in the tree. 
So for example Right Foot is the child of the Right Knee which is the child of the Right Thigh which is the child of the Hip. So the rotation of the Right Foot from the global coordinate system (centered at the hips with the $x$ axis aligned with hip joints and the $z$-axis pointing out from the hips and $y$-axis pointing upwards) is given by:
\begin{align}
    R^{\text{Tot}}_{\text{Right Foot}}=R_{\text{Right Foot}}R_{\text{Right knee}}R_{\text{Right Thigh}}R_{\text{Hip}},
\end{align}
where $R_{\text{Right Foot}},R_{\text{Right knee}},R_{\text{Right Thigh}},R_{\text{Hip}}$ are rotation matrices describing the rotation of the right foot in the right knee's coordinate system, the rotation of the right knee in the right thigh's coordinate system, the rotation of the right thigh in the hips' coordinate system. The hip's coordinate system is the global coordinate system (i.e. $R_{\text{Hip}}=I$ is identity in $\mathcal{R}^3$).
Finally, the data is a vector of exponential maps of quaternions corresponding to the local rotations (i.e. the rotation of a joint with respect to its parent such as $\text{exp}(\text{quaternion} (R_{\text{Right Foot}}))$) of all of the joints. Constantly zero-valued rotations are ignored.

\subsection{Exponential map}
\emph{Exponential map} from $\mathcal{R}^3$ to $S^3$ is defined in~\cite{Grassia98} as
\begin{align}
    e^{\mathbf v}&=[0,0,0,1] \text{   when  } \mathbf v=[0,0,0]^T\\
    e^{\mathbf v}&=\sum_{m=0}^\infty \left(\frac{1}{2}\frac{\mathbf v}{|\mathbf v|} \right)^m=\left[ \sin(\frac{1}{2}|\mathbf v|)\frac{\mathbf v}{|\mathbf v|},\cos(\frac{1}{2}|\mathbf v|)\right] \text{   when  } \boldsymbol{\mathbf v}\neq \mathbf 0.
\end{align}
Where $\mathbf v$ is a quaternion, and the multiplication in $\left(\frac{1}{2}\frac{\mathbf v}{|\mathbf v|} \right)^m$ is a quaternion multiplication.

\emph{Quaternions} $\mathbf  q=[q_x,q_y,q_z,q_w]$ consist of a vector part $ \mathbf v=[q_x,q_y,q_z]$ and a scalar part $r=q_w$.
Quaternions form a group whose underlying set is the four-dimensional vector space $\mathcal{R}^4$, with a multiplication operator that combines both the dot product and cross product of vectors [9]. The set of unit-length quaternions is a sub-group
whose underlying set is named $S^3$. Quaternions can use $S^3$ to describe and carry out rotations. The quaterion $\mathbf q=[0,0,0,1]^T$ corresponds to the identity rotation, otherwise $\mathbf q=[q_x,q_y,q_z,q_w]^T$ corresponds to rotation $\Theta=2\cos^{-1}(q_w)$ around $\hat{\mathbf v}=[q_x,q_y,q_z]$.
 A quaternion $\mathbf q$ corresponding to a rotation of $\Theta$ around $\hat{\mathbf v}\in \mathcal{R}^3$ is given by,
 \begin{align}
     \mathbf q=[q_x,q_y,q_z,q_w]=\left[\sin(\frac{1}{2}\Theta)\hat{\mathbf v},\cos(\frac{1}{2}\Theta)\right]^T.
 \end{align}
 A vector $\mathbf x \in \mathcal{R}^3$ (i.e. a quaternion with zero scale part) is rotated by a quaternion $\mathbf q$ through quaternion multiplication
  \begin{align}
     \mathbf y=\mathbf q \mathbf x\bar{\mathbf q},
 \end{align}
where $\bar{\mathbf q}$ is conjugate of $\mathbf q$ (vector part negated).
Quaternion multiplication of two quaternions $\mathbf{q}_1=\left[\mathbf{v}_1, r_1\right]$ and $\mathbf{q}_2=\left[\mathbf{v}_2, r_2\right]$ is defined as 
  \begin{align}
     \mathbf{q}_1\mathbf{q}_2=\left[r_1 \mathbf{v}_2+r_2 \mathbf{v}_1 + \mathbf{v}_1 \times\mathbf{v}_2, r_1r_2-\mathbf{v}_1 \cdot \mathbf{v}_2 \right],
 \end{align}
 where $\times$ is the cross product (i.e. vector product) of two vectors and $\cdot$ is the dot product (i.e. scalar product) of two vectors.
\end{document}